# Represented Value Function Approach for Large Scale Multi Agent Reinforcement Learning

Weiya Ren[1,2]

[1] Artificial Intelligence Research Center of National Innovation Institute of Defense Technology

[2] Tianjin Artificial Intelligence Innovation Center, Tianjin, P.R China.

E-mail: weiyren.phd@gmail.com

**Abstract** In this paper, we consider the problem of large scale multi agent reinforcement learning. Firstly, we studied the representation problem of the pairwise value function to reduce the complexity of the interactions among agents. Secondly, we adopt a l2-norm trick to ensure the trivial term of the approximated value function is bounded. Thirdly, experimental results on battle game demonstrate the effectiveness of the proposed approach.

## I. INTRODUCTION

Reinforcement learning combined with deep neural networks has achieved human-level or higher performances in challenging games [14][15]. In the recent past, deep learning has also proved to be an extremely valuable tool for multi agent reinforcement learning (MARL). The advances in deep reinforcement learning have led great interests in MARL to resolve complex and large scale problems.

One naive approach to train a MARL problem could be training all the agents using a centralized controller to obtain the joint action of all agents. However, the number of actions exponentially increases and each agent needs to send its local information to the central controller, which makes the problem intractable. [1] proposed a novel mean field method to deal with the large scale MARL problem by reducing the complexity of the interactions among agents. In contrast to [1], our contribution is twofold. Firstly, a represented value function approach is proposed to replace the mean value function to emphasize more helpful actions around each agent. In our experiments, this contribution is useful especially in the wild war scenario which all agents start with a random position and agent is insensitivity to mean actions of its neighbors. Secondly, we approximate the represented value function by a succinct function, which the omit term is ensured have a small bound.

The rest of the paper is organized as follows: Section 2 propose an new MARL approach. In Section 3 gives a brief review of related works. Experimental results are presented in Section 4. Finally, conclusions are drawn in Section 5.

## II. REPRESENTED VALUE FUNCTION

### A. Pairwise local interactions

Following [1], we factorize the Q-function using only the pairwise local interactions. Furthermore, we use a linear combination of the pairwise value functions to represent the desire value function by

$$Q^j(\boldsymbol{s},\boldsymbol{a}) = \sum_{k \in \mathcal{N}(j)} w_{j,k} Q^j(s^j, a^j, s^k, a^k). \tag{1}$$

Where $0 \leq w_{j,k} \leq 1$, $\sum_{k \in \mathcal{N}(j)} w_{j,k} = 1$, $\mathcal{N}(j)$ is the index set of the neighboring agents of agent $j$ with the size $N^j = |\mathcal{N}(j)|$.

If we denote $z^j \triangleq (s^j, a^j)$, then

$$Q^j(\boldsymbol{s},\boldsymbol{a}) = \sum_{k\in\mathcal{N}(j)} w_{j,k} Q^j(z^j, z^k). \tag{2}$$

Inspired by [1], we express the $z^k$ of each neighbor $k$ in terms of the weight sum of $z^i, i \in \mathcal{N}(j)$ and a $\delta z^{j,k}$ as:

$$z^k = z_\chi^j + \delta z^{j,k}. \tag{3}$$

where $z_\chi^j \triangleq \sum_{k\in\mathcal{N}(j)} w_{j,k} z^k$.

By Taylor's theorem, the pairwise $Q$-function $Q^j(z^j, z^k)$, if twice-differentiable with respect to $z^k$, can be expended and expressed as

$$\begin{aligned}Q^j(\boldsymbol{s},\boldsymbol{a}) &= \sum_{k\in\mathcal{N}(j)} w_{j,k} Q^j(z^j, z^k) \\ &= \sum_{k\in\mathcal{N}(j)} w_{j,k} \left[ Q^j(z^j, z_\chi^j) + \nabla_{z_\chi^j} Q^j(z^j, z_\chi^j) \delta z^{j,k} \right.\\ &\quad \left. + \frac{1}{2} \delta z^{j,k} \nabla^2_{\tilde{z}_\chi^{j,k}} Q^j(z^j, \tilde{z}_\chi^{j,k}) \delta z^{j,k} \right]. \end{aligned} \tag{4}$$

For the first term, we have

$$\sum_{k\in\mathcal{N}(j)} w_{j,k} Q^j(z^j, z_\chi^j) = Q^j(z^j, z_\chi^j). \tag{5}$$

For the second term,

$$\begin{aligned}\sum_{k\in\mathcal{N}(j)} w_{j,k} \nabla_{z_\chi^j} Q^j(z^j, z_\chi^j) \delta z^{j,k} &= \nabla_{z_\chi^j} Q^j(z^j, z_\chi^j) \left[ \sum_{k\in\mathcal{N}(j)} w_{j,k} z_\chi^j - \sum_{k\in\mathcal{N}(j)} w_{j,k} z^k \right] \\ &= \nabla_{z_\chi^j} Q^j(z^j, z_\chi^j)[z_\chi^j - z_\chi^j] = 0. \end{aligned} \tag{6}$$

where $\tilde{z}_\chi^{j,k} = z_\chi^j + \epsilon^{j,k} \delta z^{j,k}$, $\epsilon^{j,k} \in [0,1]$.

Importantly, l2-norm operation for $s^j, a^j, \sum_{k\in\mathcal{N}(j)} w_{j,k} s^k, \sum_{k\in\mathcal{N}(j)} w_{j,k} a^k$ (denote as $\underline{s^j}, \underline{a^j}, \underline{\sum_{k\in\mathcal{N}(j)} w_{j,k} s^k}, \underline{\sum_{k\in\mathcal{N}(j)} w_{j,k} a^k}$), $\left\|\sum_{k\in\mathcal{N}(j)} w_{j,k} s^k\right\| = 1, \left\|\sum_{k\in\mathcal{N}(j)} w_{j,k} a^k\right\| = 1, \|\underline{s^j}\| = 1, \|\underline{a^j}\| = 1$) can ensure $\sum_{k\in\mathcal{N}(j)} w_{j,k} \frac{1}{2} \delta z^{j,k} \nabla^2_{\tilde{z}_\chi^{j,k}} Q^j(z^j, \tilde{z}_\chi^{j,k}) \delta z^{j,k}$ is bounded within a symmetric interval $[-4M, 4M]$ under the mild condition of the $Q$-function $Q^j(z^j, z^k)$ being $M$-smooth.

Thus, with sharing representation weights, the value function of agent $j$ is approximated by

$$Q^j(\boldsymbol{s},\boldsymbol{a}) \approx Q^j\left(\underline{s^j}, \underline{a^j}, \underline{\sum_{k\in\mathcal{N}(j)} w_{j,k} s^k}, \underline{\sum_{k\in\mathcal{N}(j)} w_{j,k} a^k}\right). \tag{7}$$

## B. Implementation

Following [1], for off-policy learning, we exploit standard Q-learning [5] for discrete action spaces, which we call RFQ (represented value function for Q-learning). Agent $j$ is trained by

minimizing the loss function

$$\mathcal{L}(\phi^j) = (y^j - Q_{\phi^j}(\underline{s^j}, \underline{a^j}, \underbrace{\sum_{k \in \mathcal{N}(j)} w_{j,k} s^k}, \underbrace{\sum_{k \in \mathcal{N}(j)} w_{j,k} a^k}))^2. \quad (8)$$

where $y^j = r^j + \gamma v_{\phi_-^j}(s')$ is the target mean field value calculated with the weights $\phi_-^j$. Differentiating $\mathcal{L}(\phi^j)$ gives

$$\nabla_{\phi^j} \mathcal{L}(\phi^j)$$
$$= (y^j - Q_{\phi^j}(\underline{s^j}, \underline{a^j}, \underbrace{\sum_{k \in \mathcal{N}(j)} w_{j,k} s^k}, \underbrace{\sum_{k \in \mathcal{N}(j)} w_{j,k} a^k})) \nabla_{\phi^j} Q_{\phi^j}(\underline{s^j}, \underline{a^j}, \underbrace{\sum_{k \in \mathcal{N}(j)} w_{j,k} s^k}, \underbrace{\sum_{k \in \mathcal{N}(j)} w_{j,k} a^k}). \quad (9)$$

The new Boltzmann policy is then determined for each $j$ that

$$\pi_t^j\left(a^j \middle| \underline{s^j}, \underline{a^j}, \underbrace{\sum_{k \in \mathcal{N}(j)} w_{j,k} s^k}, \underbrace{\sum_{k \in \mathcal{N}(j)} w_{j,k} a^k}\right) = \frac{exp(-\beta Q_t^j(\underline{s^j}, \underline{a^j}, \underline{\sum_{k \in \mathcal{N}(j)} w_{j,k} s^k}, \underline{\sum_{k \in \mathcal{N}(j)} w_{j,k} a^k}))}{\sum exp(-\beta Q_{\phi^j}(\underline{s^j}, \underline{a^j}, \underline{\sum_{k \in \mathcal{N}(j)} w_{j,k} s^k}, \underline{\sum_{k \in \mathcal{N}(j)} w_{j,k} a^k}))}. \quad (10)$$

Following [1], instead of setting up Boltzmann policy using the Q-function as in RFQ, we can explicitly model the policy by neural networks with the weights $\theta$, which leads to the on-policy actor-critic method [4] that we call RFAC. The policy network $\pi_{\theta^j}$ is trained by the sampled policy gradient

$$\nabla_{\theta^j} \mathcal{J}(\theta^j) \approx \nabla_{\theta^j} log \pi_{\theta^j}(s) Q_{\phi^j}(\underline{s^j}, \underline{a^j}, \underbrace{\sum_{k \in \mathcal{N}(j)} w_{j,k} s^k}, \underbrace{\sum_{k \in \mathcal{N}(j)} w_{j,k} a^k})|_{a = \pi_{\theta^j}(s)}. \quad (11)$$

During the training of MF-AC, one needs to alternatively update $\phi$ and $\theta$ until convergence.

The representation weights $w$ can be defined in various way. In this paper, we adopt the GAT method [6] to compute the weights.

## III. RELATED WORK

Literature in MARL has argued for a centralized training decentralized execution scheme, e.g., MADDPG [10], BICNET[11] and COMA[12]. Naturally, such a scheme is not scalable as the input size of the critic network grows. As the number of agents grows larger, the above methods are no longer feasible. Another scheme is to approximate the policies of all the agents using meta-learning [13], which is also infeasible when facing a large number of agents. The work in [1] proposed a mean field based approach that take the mean actions of neighbors of each agent into consideration, which can deal with the MARL problem with a large number of agents.

We notice that the work in [7](MAAC) is closest to our work, where the authors use centrally computed critics that share an attention mechanism which selects relevant information for each agent. Our work is inspired by work in [1] which each agent takes its neighbor's actions into primary consideration. Furthermore, the proposed approach aims to the solve the MARL problem with hundreds of agents. The work in [8] utilize a graph convolutional approach [9] to extract features of samples, while our work also utilize the graph convolutional approach to compute the representation weights.

## IV. EXPERIMENT RESULTS

In this section, we analyze and evaluate our algorithms in the Open-source MAgent system

[2], which is a mixed cooperative-competitive battle game with two armies fighting against each other in a grid world, each empowered by a different multi agent reinforcement learning (MARL) algorithm.

*A. Environment*

For fair comparison, we adopt the default settings for training Battle game in all experiments (same with [1]): −0.005 for every move, 0.2 for attacking an enemy, 5 for killing an enemy, −0.1 for attacking an empty grid, −0.1 for being attacked or killed, 64 homogeneous agents for each army, 400 iterations for each round and 2000 rounds self-plays without any knowledge of other MARL algorithms. Following [2], the goal of each army is to get more rewards for training and destroy all the opponents in testing (that's why we do not change the default reward settings to encourage attack).

*B. Model Settings*

Our RFQ and RFAC are compared against the baselines that are proved successful on the MAgent platform. We focus on the battles between the proposed algorithms and the mean field methods (MFQ, MFAC) [1], independent Q-learning (IL) and advantageous actor critic (AC). Following [1], we exclude MADDPG/MAAC as baselines, as they cannot deal with the varying number of agents for the battle. Also, [1] demonstrated that MAAC tends to scale poorly and fail when the agent number is in hundreds.

*C. Results and Discussion*

Following all parameter settings in [1], we train all six models in the default battle scenario. For testing, we use them for comparative games in two scenarios, which are the default battle scenario and the wild war scenario. The unique difference between them is that agents start with two divided camps in the default battle scenario and agents start with a random position in the wild war scenario, as shown in Fig. 1. Training in the default battle scenario and testing in both the default battle scenario and the wild war scenario can test the generalization ability of models. Some game silhouettes are shown in Fig. 2.

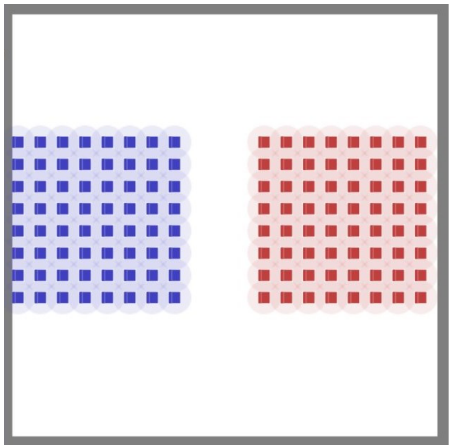
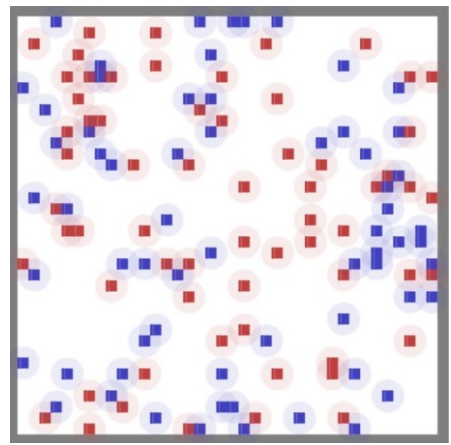

(a) Start with two divided camps (Battle scenario).      (b) Start with random positions (Wild war scenario).

Fig. 1 First frame of the Battle scenario and the Wild war scenario.

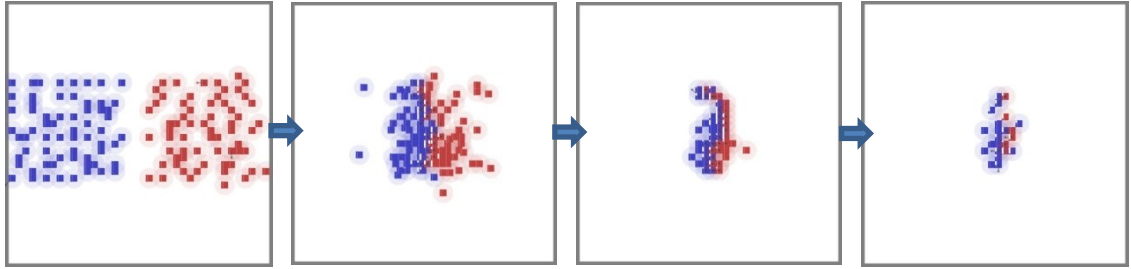

(a)Progress of a game in the Battle scenario. (RFAC vs MFAC)

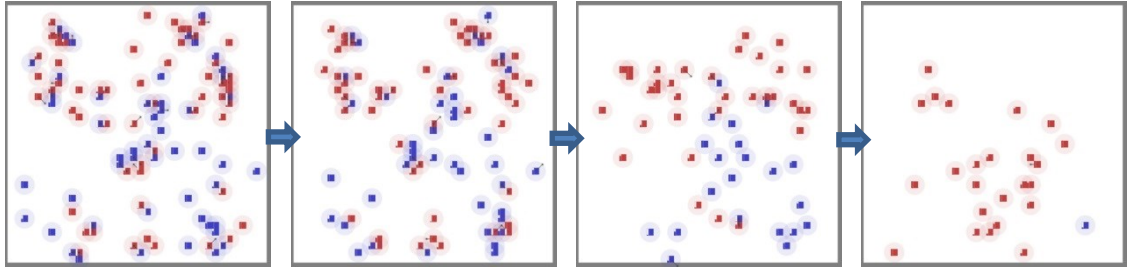

(b)Progress of a game in the Wild war scenario. (RFAC vs MFAC)

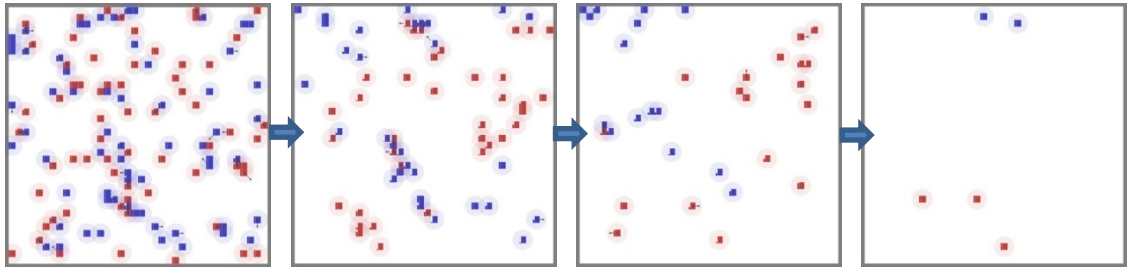

(c)Progress of a game in the Wild war scenario. (RFAC vs RFAC)

Fig. 2 Progress of some games.

In training phase, each algorithm is trained three times independently. After training, 18 models are created as 18 independent players (IL_A, IL_B, IL_C, MFQ_A, MFQ _B, MFQ _C, RFQ_A, RFQ_B, RFQ_C, AC_A, AC_B, AC_C, MFAC_A, MFAC_B, MFAC_C, RFAC_A, RFAC_B, RFAC_C). In testing phase, skill levels of all players are estimated by calculating Elo scores (adapted from chess [3]) based on outcomes of battle or wild war. For each testing game, we randomly select two players among the 18 players to play.

For battle scenario, we played about 136,000 games in total. The result demonstrate that actor-critic MARL algorithms outperform Q-learning MARL algorithms, as result shown in Tab. 1. The proposed RFAC algorithm performs well while the proposed RFQ algorithm has a general performance in battle scenario.

As shown in Tab. 2, the proposed RFQ algorithm and MFAC nearly went to deuce, however, the actor-critic MARL algorithms swept RFQ. As shown in Tab. 3, the proposed RFAC algorithm has a good performance when fight with the other algorithms.

Tab. 1 Testing result in Battle scenario

| Rank | Players | Elo scores | Players | Kills/Be-killed ratio | Kills | Players | Winrate | Played games | Draws |
|---|---|---|---|---|---|---|---|---|---|
| 1 | RFAC_C | 1720.91 | RFAC_C | 1.3362 | 927,015 | RFAC_C | 75.08% | 15,062 | 54 |
| 2 | RFAC_A | 1706.36 | RFAC_A | 1.3316 | 930,819 | RFAC_A | 74.74% | 15,130 | 53 |
| 3 | AC_A | 1650.09 | RFAC_B | 1.3307 | 928,920 | RFAC_B | 74.41% | 15,105 | 72 |
| 4 | AC_B | 1642.14 | AC_A | 1.2699 | 917,830 | AC_B | 72.16% | 15,189 | 126 |
| 5 | MFAC_B | 1640.15 | AC_B | 1.2667 | 913,396 | AC_A | 72.03% | 15,258 | 120 |
| 6 | MFAC_C | 1637.71 | AC_C | 1.2635 | 921,287 | AC_C | 71.36% | 15,342 | 120 |
| 7 | RFAC_B | 1616.80 | MFAC_A | 1.2408 | 911,124 | MFAC_C | 69.29% | 15,287 | 204 |
| 8 | AC_C | 1591.63 | MFAC_C | 1.2384 | 924,616 | MFAC_A | 69.14% | 15,066 | 207 |
| 9 | MFAC_A | 1572.06 | MFAC_B | 1.2366 | 922,698 | MFAC_B | 68.78% | 15,254 | 219 |
| 10 | MFQ_C | 1234.99 | RFQ_C | 0.8153 | 736,159 | MFQ_C | 36.09% | 14,931 | 72 |
| 11 | MFQ_A | 1217.33 | RFQ_B | 0.8147 | 738,367 | MFQ_B | 35.73% | 15,201 | 88 |
| 12 | MFQ_B | 1199.96 | RFQ_A | 0.8131 | 728,077 | MFQ_A | 35.10% | 15,156 | 87 |
| 13 | RFQ_A | 1159.45 | MFQ_C | 0.7836 | 646,796 | RFQ_C | 25.62% | 15,190 | 64 |
| 14 | IL_A | 1148.05 | MFQ_B | 0.7787 | 655,983 | RFQ_B | 25.50% | 15,265 | 74 |
| 15 | RFQ_B | 1126.04 | MFQ_A | 0.7759 | 651,414 | RFQ_A | 25.38% | 15,061 | 71 |
| 16 | IL_B | 1121.24 | IL_C | 0.7177 | 639,727 | IL_C | 21.08% | 15,281 | 62 |
| 17 | IL_C | 1120.04 | IL_B | 0.7159 | 625,719 | IL_A | 21.05% | 15,185 | 63 |
| 18 | RFQ_C | 1095.06 | IL_A | 0.7158 | 633,811 | IL_B | 21.01% | 14,985 | 66 |

Tab. 2 Wins contrast when RFQ fight against IL,MFQ,AC,MFAC and RFAC.

| Belligerents | IL(A+B+C) | MFQ(A+B+C) | AC(A+B+C) | MFAC(A+B+C) | RFAC(A+B+C) |
|---|---|---|---|---|---|
| RFQ(A+B+C) | 4887:2889 | 3983:3941 | 1 : 8105 | 1 : 8056 | 0 : 7972 |

Tab. 3 Wins contrast when RFAC fight against IL,MFQ,RFQ,AC and MFAC.

| Belligerents | IL(A+B+C) | MFQ(A+B+C) | RFQ(A+B+C) | AC(A+B+C) | MFAC(A+B+C) |
|---|---|---|---|---|---|
| RFAC(A+B+C) | 8000 : 0 | 5420 : 2465 | 7972 : 0 | 4295:3765 | 5625:2493 |

For wild war scenario, we played about 100,000 games in total. The result is shown in Tab. 4. It is evident that on all the metrics the proposed RFAC largely outperforms the corresponding baselines, i.e. IL, MFQ, AC and MFAC respectively. The proposed RFQ algorithm largely outperforms IL and MFQ, respectively. The mean filed approach preforms not good in the wild war scenario which is insensitivity to mean actions.

Tab. 4 Testing result in Wild war scenario

| Rank | Players | Elo scores | Players | Kill/Be-killed ratio | Kills | Players | Winrate | Played games | Draws |
|---|---|---|---|---|---|---|---|---|---|
| 1 | RFAC_C | 2953.19 | RFAC_C | 1.9931 | 763,071 | RFAC_C | 93.93% | 12,109 | 12 |
| 2 | RFAC_B | 2929.04 | RFAC_B | 1.9886 | 754,882 | RFAC_B | 93.92% | 11,981 | 10 |
| 3 | RFAC_A | 2820.07 | RFAC_A | 1.9832 | 748,070 | RFAC_A | 93.72% | 11,875 | 12 |
| 4 | RFQ_B | 1822.49 | RFQ_C | 1.1245 | 655,094 | RFQ_C | 69.78% | 11,994 | 155 |
| 5 | RFQ_A | 1822.09 | RFQ_B | 1.1238 | 652,306 | RFQ_B | 69.56% | 11,946 | 166 |
| 6 | RFQ_C | 1736.07 | AC_C | 1.1215 | 533,157 | RFQ_A | 68.65% | 11,987 | 154 |
| 7 | AC_A | 1619.43 | AC_A | 1.1199 | 531,259 | AC_C | 58.16% | 11,784 | 264 |
| 8 | AC_C | 1617.97 | RFQ_A | 1.1124 | 651,599 | AC_A | 57.80% | 11,733 | 242 |
| 9 | AC_B | 1575.58 | AC_B | 1.1119 | 531,079 | AC_B | 56.71% | 11,715 | 257 |
| 10 | MFAC_B | 1446.66 | MFAC_A | 1.0100 | 502,233 | MFAC_C | 46.31% | 11,844 | 232 |
| 11 | MFAC_A | 1426.28 | MFAC_C | 1.0100 | 498,845 | MFAC_A | 45.98% | 11,965 | 235 |
| 12 | MFAC_C | 1372.44 | MFAC_B | 1.0041 | 487,220 | MFAC_B | 45.75% | 11,602 | 209 |
| 13 | IL_A | 638.80 | IL_B | 0.6434 | 422,307 | IL_A | 22.71% | 11,680 | 57 |
| 14 | IL_C | 629.33 | IL_C | 0.6433 | 423,934 | IL_B | 22.55% | 11,787 | 55 |
| 15 | IL_B | 556.73 | IL_A | 0.6432 | 418,409 | IL_C | 22.54% | 11,835 | 64 |
| 16 | MFQ_B | 132.83 | MFQ_B | 0.5287 | 304,149 | MFQ_B | 6.79% | 11,874 | 20 |
| 17 | MFQ_A | 80.80 | MFQ_C | 0.5247 | 300,540 | MFQ_C | 6.58% | 11,802 | 19 |
| 18 | MFQ_C | 20.22 | MFQ_A | 0.5234 | 298,666 | MFQ_A | 6.52% | 11,803 | 25 |

As shown in Tab. 5, the proposed RFQ algorithm outperforms all the other algorithms except RFAC. As shown in Tab. 6, the proposed RFAC algorithm has far superior forces with contrast to the other algorithms.

Tab. 5 Wins contrast when RFQ fight against IL,MFQ,AC,MFAC and RFAC

| Belligerents | IL(A+B+C) | MFQ(A+B+C) | AC(A+B+C) | MFAC(A+B+C) | RFAC(A+B+C) |
|---|---|---|---|---|---|
| RFQ(A+B+C) | 6335:5 | 6217:7 | 4545:1576 | 5714:461 | 0:6398 |

Tab. 6 Wins contrast when RFAC fight against IL,MFQ,RFQ,AC and MFAC.

| Belligerents | IL(A+B+C) | MFQ(A+B+C) | RFQ(A+B+C) | AC(A+B+C) | MFAC(A+B+C) |
|---|---|---|---|---|---|
| RFAC(A+B+C) | 6301:0 | 6314:0 | 6398:0 | 6186:0 | 6380:0 |

The win-table among all the 18 players of the two scenarios is shown in Fig. 3, we can find that the actor-critic algorithms always have a good performance. Furthermore, the proposed algorithms have better performance in the wild war scenario.

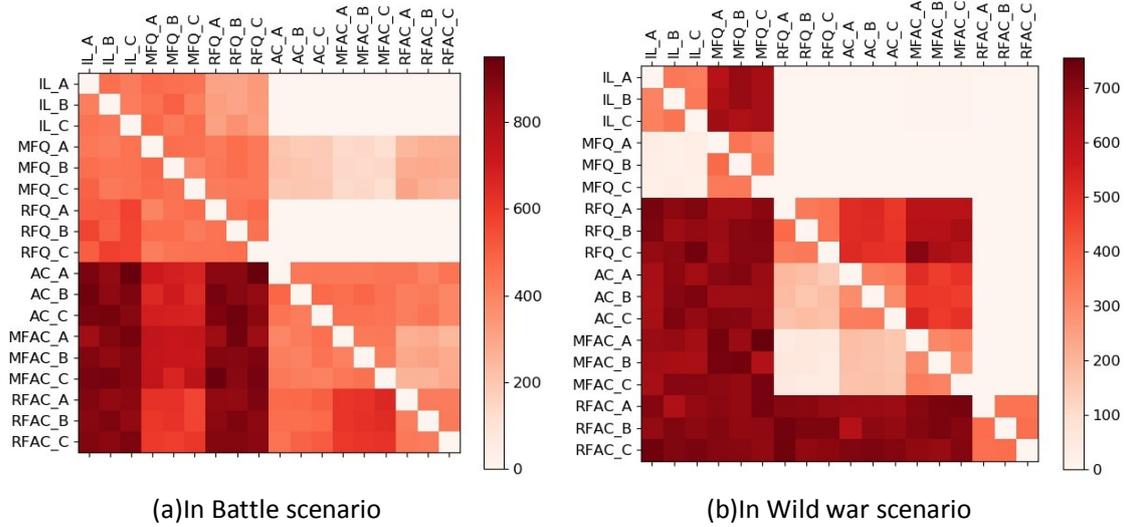

(a)In Battle scenario  (b)In Wild war scenario

Fig. 3 Win-table with all the belligerents. The line of the win-table means the number of wins of all the games between the corresponding player and the other players(17 players), respectively.

## V. CONCLUSION AND FUTURE WORK

In this paper, we proposed a novel method for large scale MARL problem, which named RFQ and RFAC. Firstly, we studied the weight norm of the value function. Secondly, we ensured the approximated value function's trivial term is bounded. Thirdly, the proposed algorithms for battle game have a good performance especially on the wild war scenario, which shows the robustness of the proposed algorithms. More approaches to compute the representation weights and more applications will to be investigated in our future work. The code of this paper is at https://github.com/renweiya/RFQ-RFAC.